\documentclass[sigconf]{acmart}
\usepackage{makecell}

\AtBeginDocument{%
  }

\begin{document}

\title{Text Classification: Neural Networks VS Machine Learning Models VS Pre-trained Models}

\author{Christos Petridis}
\email{christos.petridis@temple.edu}
\orcid{0009-0002-2535-2179}
\affiliation{%
  \institution{Temple University}
  \city{Philadelphia}
  \state{Pennsylvania}
  \country{USA}
}

\renewcommand{\shortauthors}{Petridis et al.}

\begin{abstract}
Text classification is a very common task nowadays and there are many efficient methods and algorithms that we can employ to accomplish it. Transformers have revolutionized the field of deep learning, particularly in Natural Language Processing (NLP) and have rapidly expanded to other domains  such as computer vision \cite{jamil2023comprehensive} \cite{steiner2024paligemma}, time-series analysis \cite{ahmed2023transformers}, and more. The transformer model was firstly introduced in the context of machine translation and its architecture relies on self-attention mechanisms \cite{vaswani2017attention} to capture complex relationships within data sequences. It is able to handle long-range dependencies more effectively than traditional neural networks (such as Recurrent Neural Networks and Multilayer Perceptrons). In this work, we present a comparison  between different techniques to perform text classification. We take into consideration seven pre-trained models, three standard neural networks and three machine learning models. For standard neural networks and machine learning models we also compare two embedding techniques: TF-IDF and GloVe, with the latter consistently outperforming the former. Finally, we demonstrate the results from our experiments where pre-trained models such as BERT and DistilBERT always perform better than standard models/algorithms. 
\end{abstract}



\keywords{Text Classification, Transformers, Pre-Trained Models, Natural Language Processing, Embeddings}


\maketitle

\section{Introduction}
NLP has been a rapidly growing area due to its impact on how humans interact with technology. We have seen many applications ranging from voice-activated assistants and chatbots to information retrieval, text summarization, and sentiment analysis. Its relevance continues to expand with advances in GenAI (Generative AI), making it a field with vast research potential and numerous practical applications. NLP continues to expand due to its great capabilities in understanding, generating, and transforming human language. Some popular examples are the great advances in Large Language Models (LLMs) like ChatGPT (models: GPT-3.5, GPT-4o, GPT-4o mini, GPT-4 Turbo, GPT-4 etc.), LLaMA (models: Llama 3.1, Llama 3.2 etc.) and others. These models have revolutionized the field by achieving levels of fluency, comprehension, and contextual understanding like never before. LLMs have transformed the field of NLP with their ability to learn patterns in text from massive datasets. We have seen results that were not possible with earlier rule-based models. LLMs can generate human-like text, understand complex queries, and even perform tasks on which they were not explicitly trained, thanks to their pre-trained and transfer learning capabilities.

Computers and, of course, AI models understand and generate human language through the use of embeddings. Embeddings in NLP refer to dense vector representations of words, phrases, or even entire sentences that capture semantic meaning and contextual relationships within a numerical format that models can understand. Unlike traditional one-hot encoding methods, which are very sparse and fail to convey any information about how words relate to each other, embeddings encode words into continuous, low-dimensional spaces. This allows models to map semantically similar words close to each other in this space. For example, the words "king" and "queen" may appear closer to each other in the vector space than "king" and "car" because of their semantic and contextual similarities. Embeddings are learned by models based on large corpora of text, making them effective at capturing intricate relationships, patterns, and linguistic properties. Famous embedding techniques such as Word2Vec, GloVe (Global Vectors) and contextual embeddings from transformer-based models represent a word in context, considering the surrounding words in a sentence. This means that the word "bank" would have different embeddings depending on whether it is used in the context of a riverbank or a financial institution. It is clear that those embeddings offer better results in NLP tasks than older static embeddings such as TF-IDF (Term Frequency-Inverse Document Frequency).

In this work, we present a performance comparison between pre-trained models and standard models on a classification task. We have developed three neural networks for our experiments including MLP, RNN, and TransformerEncoder models and we refer to them as standard models. We will also present the performance of some machine learning models (Support Vector Machines, Random Forest and Logistic Regression) and we refer to them as machine learning models. We will show the performance of both the standard and machine learning models employing different embeddings (TF-IDF and GloVe). Regarding the pre-trained models, we will use their corresponding embeddings since each model has its own predefined embeddings. The structure of this work is as follows: Section \ref{sec:relatedwork} reviews relevant works, establishing the context for our work. Section \ref{sec:embeddings} introduces the embedding techniques used, Section \ref{sec:dataset} describes the dataset and Section \ref{sec:datapreprocessing} outlines data pre-processing steps. Sections \ref{sec:neuralnets} and \ref{sec:mlmodels} cover the neural network architectures and machine learning models applied. Section \ref{sec:transferlearning} highlights the use of transfer learning to enhance performance. Results are presented in Section \ref{sec:results}, followed by a discussion in Section \ref{sec:discussion} and a conclusion with key findings in Section \ref{sec:conclusions}.

\section{Related Work}
\label{sec:relatedwork}
We have seen many surveys in the past regarding text classification methods. In \cite{kowsari2019text} a wide survey is presented on different classification algorithms and embeddings. They present results from other papers with no explicit comparison between embeddings and models. In \cite{minaee2021deep} authors demonstrate a deep learning-based review where they experiment with various text datasets such as AG News, 20 Newsgroups, Reuters news and many more. Moreover, \cite{ExploringTransformersmodels} shows a comparison of BERT, DistilBERT, RoBERTa, XLNET and ELECTRA for emotion recognition where they used BERT as their baseline model and compared it with the four additional transformer-based models (DistillBERT, RoBERTa, XLNet, and ELECTRA). Except for the ELECTRA model, which had the worst F1-score (.33), the other models had very similar results. RoBERTa achieved the best F1-score (.49), followed by DistillBERT (.48), XLNet (.48), and then BERT (.46). Inspired by all these works, we are presenting a comparison between various pre-trained models, neural networks and standard models employing two different techniques for embeddings (TF-IDF and GloVe).

\section{Embeddings}
\label{sec:embeddings}
Embeddings are a crucial aspect of how computers understand and generate human language. They represent words or phrases as vectors in a high-dimensional space, capturing semantic relationships and contextual meanings. In this section, we will explore two techniques for generating embeddings: TF-IDF \cite{tfidf} and GloVe \cite{glove}.

Term Frequency-Inverse Document Frequency (TF-IDF) is a statistical method used to evaluate the importance of a word in a document relative to a collection of documents (corpus). It is calculated as the product of two components: Term Frequency (TF) and Inverse Document Frequency (IDF). The term frequency is defined as:
\begin{equation}
\text{TF}(t, d) = \frac{f_{t, d}}{\sum_{t' \in d} f_{t', d}},
\end{equation}
where \(f_{t, d}\) is the frequency of term \(t\) in document \(d\), and the denominator is the total number of terms in \(d\).

The inverse document frequency is defined as:
\begin{equation}
\text{IDF}(t, D) = \log \frac{|D|}{|\{d \in D : t \in d\}|}
\end{equation}
where \(|D|\) is the total number of documents in the corpus, and \(|\{d \in D : t \in d\}|\) is the number of documents containing term \(t\).

The TF-IDF score is then computed as:
\begin{equation}
\text{TF-IDF}(t, d, D) = \text{TF}(t, d) \cdot \text{IDF}(t, D)
\end{equation}

TF-IDF focuses on capturing the importance of terms by penalizing frequently occurring terms across the corpus, ensuring that common words like "the" or "and" are not overemphasized.

Global Vectors for Word Representation (GloVe) is a popular unsupervised learning algorithm for generating word embeddings. Unlike TF-IDF, GloVe captures the semantic relationships between words by analyzing word co-occurrence statistics in a corpus. In this work, we use the pre-trained GloVe embeddings from the \texttt{fse/glove-wiki-gigaword-100} dataset, which provides 100-d vectors trained on Wikipedia and Gigaword data. These embeddings encode rich semantic and syntactic relationships, such as analogies and word similarities. While TF-IDF emphasizes term importance in individual documents, GloVe captures broader semantic relationships from the entire corpus.

\section{Dataset}
\label{sec:dataset}
The dataset we used to implement the experiments is a news datasets which contains 10917 examples/articles, 11 features and 2 classes for each example (13 columns in total). Table \ref{tab:dataset_description} demonstrates some details regarding the dataset and briely describes each column. However, identification (such as \texttt{data\_id} and \texttt{id}), timestamps and temporal information do not help as much as the content itself  when trying to classify articles into categories. Therefore, we will use only 4 features: \texttt{source, title, content} and \texttt{author}.
\begin{table}[h]
\caption{The initial dataset}
\begin{tabular}{p{0.3\columnwidth}p{0.6\columnwidth}}
\toprule
\textbf{Column Name} & \textbf{Description} \\
\midrule
\texttt{data\_id} & Unique identifier number for the article. \\
\texttt{id} & source + date + article \\
\texttt{date} & Date associated with the entry. \\
\texttt{source} & Source of the article. \\
\texttt{title} & Title of the article. \\
\texttt{content} & Main content of the article. \\
\texttt{author} & Author of the article. \\
\texttt{url} & URL link to the article. \\
\texttt{published} & Date and time of publication. \\
\texttt{published\_utc} & The unix timestamp of the publication. \\
\texttt{collection\_utc} & The unix timestamp of the time the incident recorded. \\
\texttt{category\_level\_1} & Level-1 category. \\
\texttt{category\_level\_2} & Level-2 category. \\
\bottomrule
\end{tabular}
\label{tab:dataset_description}
\end{table}

\subsection{Distribution of classes}
\begin{figure}[h]
  \centering
  \includegraphics[width=\linewidth]{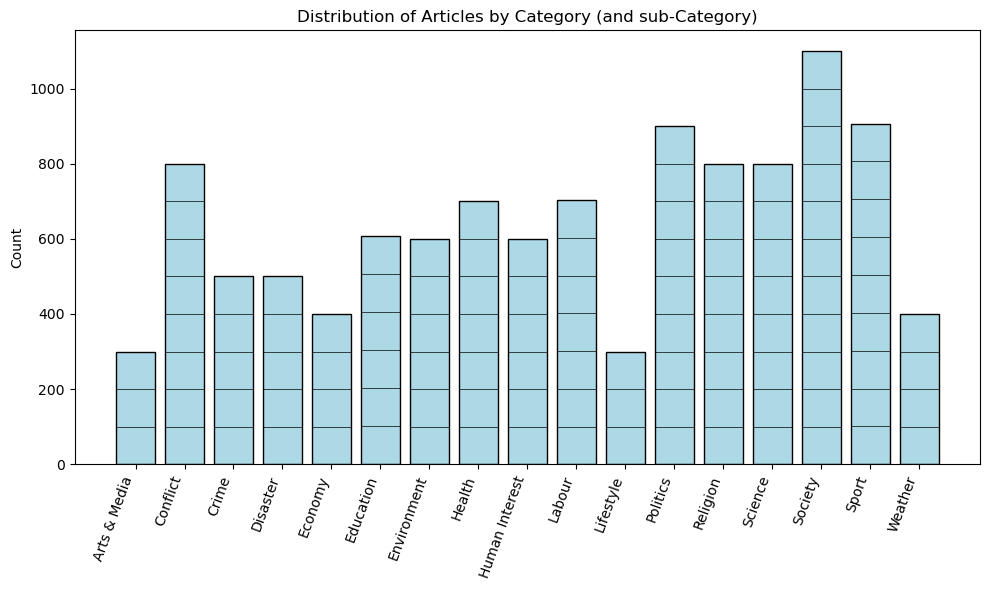}
  \caption{Distribution of the classes in our dataset. It is also evident how many level-2 categories we have under each level-1 category.}
  \label{fig:datasetclasses}
\end{figure}

The distribution of classes in the dataset is shown in Figure \ref{fig:datasetclasses}. We have 17 level-1 categories and 109 level-2 categories. Under each level-1 category we have approximately 100 instances for each level-2 category. Broadly speaking, we can say that it is a balanced dataset because we always have 100 instances for a combination of level-1 and level-2 categories.

\section{Data pre-processing}
\label{sec:datapreprocessing}
This step is very crucial in tasks that involve NLP. The goal is to transform raw text data into a format that can be used by (machine/deep learning) models \cite{ranjan2016survey}, \cite{sun2014empirical}.
\subsection{Merge the Features}
In NLP we have to merge all the information (e.g., features) we have into one feature/column. Therefore, before the actual data preprocessing we have to merge \texttt{source, title, content} and \texttt{author} into one column. Therefore, the final dataset has 3 columns: the information of each article, level-1 and level-2 categories.
\subsection{Data Cleaning and Tokenization}
Having all the information in a single feature, the first thing that we need to do is to convert all text to lowercase for consistency, so words with different cases are treated the same. Then, we have to remove numbers, URLs, special characters and punctuation that add noise to the data and do not contribute to our task.
\subsection{Lemmatization and Stopwords Removal}
This step transforms words into their base or dictionary forms, which helps group similar words together and reduces the overall vocabulary size. For example, different forms of a word like "going" or "went" are reduced to their root form "go". Additionally, commonly used words, known as stopwords (such as "the", "a", "and", "is"), are removed since they do not usually contribute significant meaning to the context. All of the above are performed using functions such as WordNetLemmatizer and the stopwords from the NLTK library. After this stage, we are ready for tokenization or embeddings.

\section{Neural Networks}
\label{sec:neuralnets}
Since we are using the 100-d vectors for GloVe embeddings and also 100 for max features in TF-IDF embeddings, the input vector for the neural networks is going to be 100.

\begin{itemize}
    \item \textbf{MLP:} A feedforward neural network \cite{taud2018multilayer} which consists of two hidden layers (256 and then 128 neurons) with ReLU activation functions, each followed by a dropout layer to reduce overfitting by randomly setting 20\% of activations to zero during training. The final layer outputs predictions for 17 (level-1 category) or 109 (level-2 category) classes using a linear transformation.
    
    \item \textbf{Transformer:} A transformer-based architecture \cite{vaswani2017attention} which begins by applying a linear embedding to the input data to map it to a 64-d space. The 64-d data is then processed through a stack of Transformer Encoder layers, which use self-attention mechanisms to capture relationships between input features. After that, the output is being processed by taking the mean across the sequence dimensions, resulting in a single vector representation. After that, the single vector is passed through a fully connected layer to predict the class probabilities for the specified number of classes (17 or 109). We also have applied dropout (20\%) during embedding and within the Transformer layers to reduce overfitting.
    
    \item \textbf{RNN:} A recurrent neural network \cite{medsker2001recurrent} which includes an RNN layer with 64 hidden units and 1 layer, followed by a fully connected layer to map the last hidden state to the output dimension (17 or 109). During the forward pass, the input sequence is reshaped to match the expected dimensions, and only the output from the final time step is passed to the fully connected layer.
\end{itemize}

\subsection{Training Phase}
\sloppy
For the training phase we employ the Adam optimizer and CrossEntropy loss function over 150 epochs. We apply the torch.optim.lr\_scheduler.ReduceLROnPlateau as the learning rate scheduler, which dynamically adjusts the learning rate based on the model's performance on the test set, promoting stable convergence. We start with LR=0.001 and it decreases based on the performance (minimum LR=1e-5). Depending on availability, we use either GPU or CPU, with the best model state saved for further evaluation. In our case, we were fortunate enough to have access in a Tesla V100-SXM2-16GB GPU.

\subsection{Neural Networks Results}
\begin{figure}[h]
  \centering
  \includegraphics[width=\linewidth]{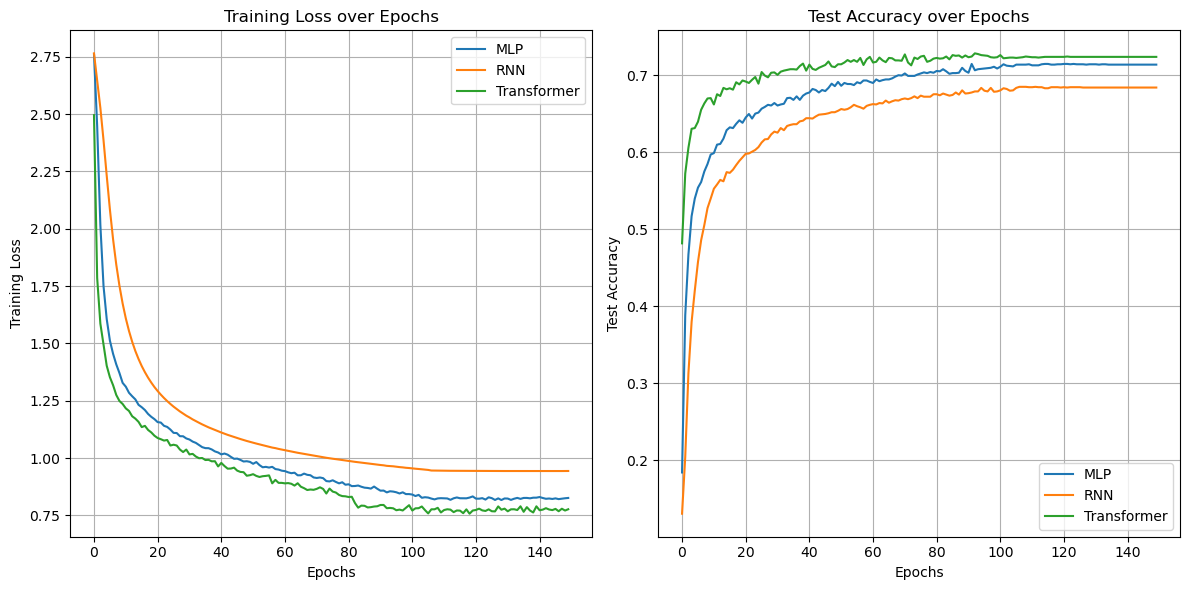}
  \caption{Training Loss (on the left) and Test Accuracy (on the right) employing GloVe for level-1 category.}
  \label{fig:glove1}
\end{figure}

\begin{figure}[h]
  \centering
  \includegraphics[width=\linewidth]{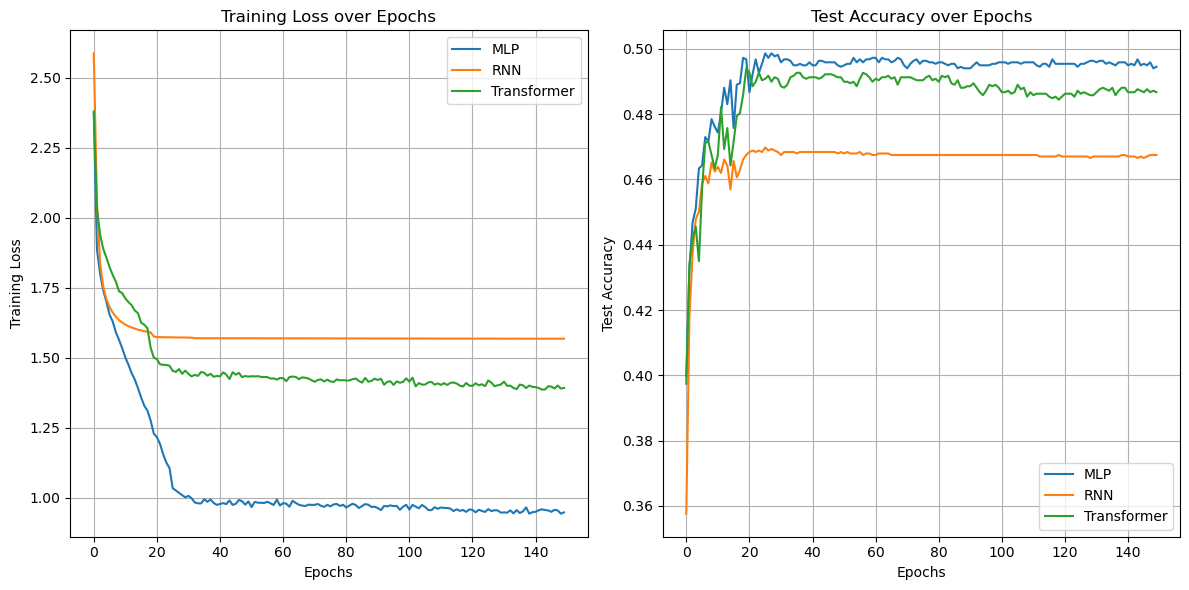}
  \caption{Training Loss (on the left) and Test Accuracy (on the right) employing TF-IDF for level-1 category.}
  \label{fig:tfidf1}
\end{figure}

\begin{figure}[h]
  \centering
  \includegraphics[width=\linewidth]{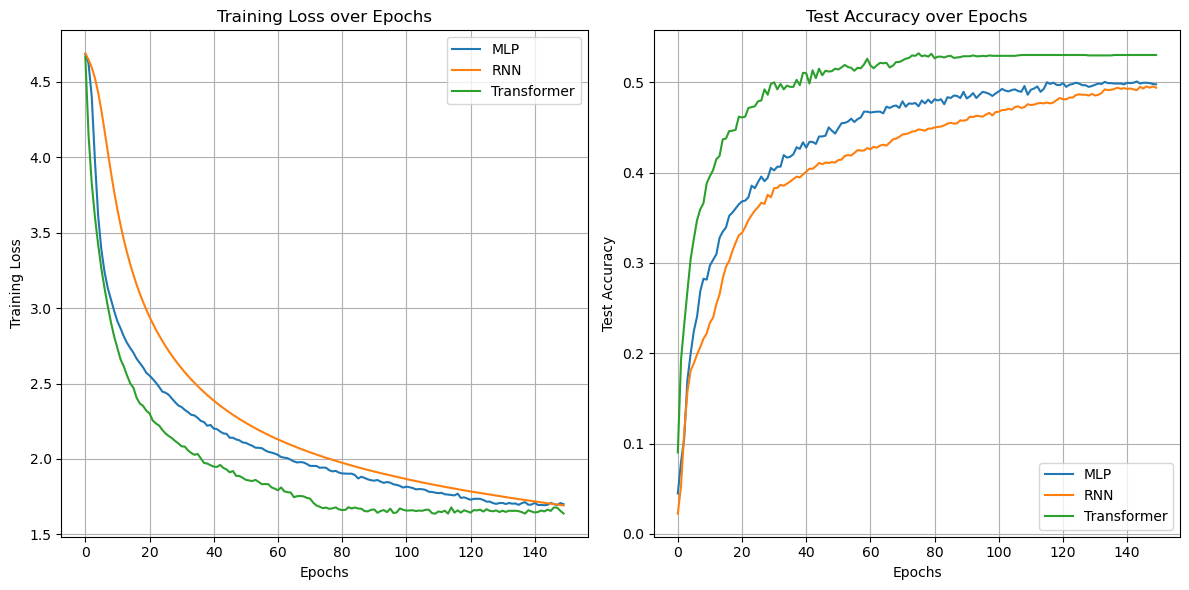}
  \caption{Training Loss (on the left) and Test Accuracy (on the right) employing GloVe for level-2 category.}
  \label{fig:glove2}
\end{figure}

\begin{figure}[h]
  \centering
  \includegraphics[width=\linewidth]{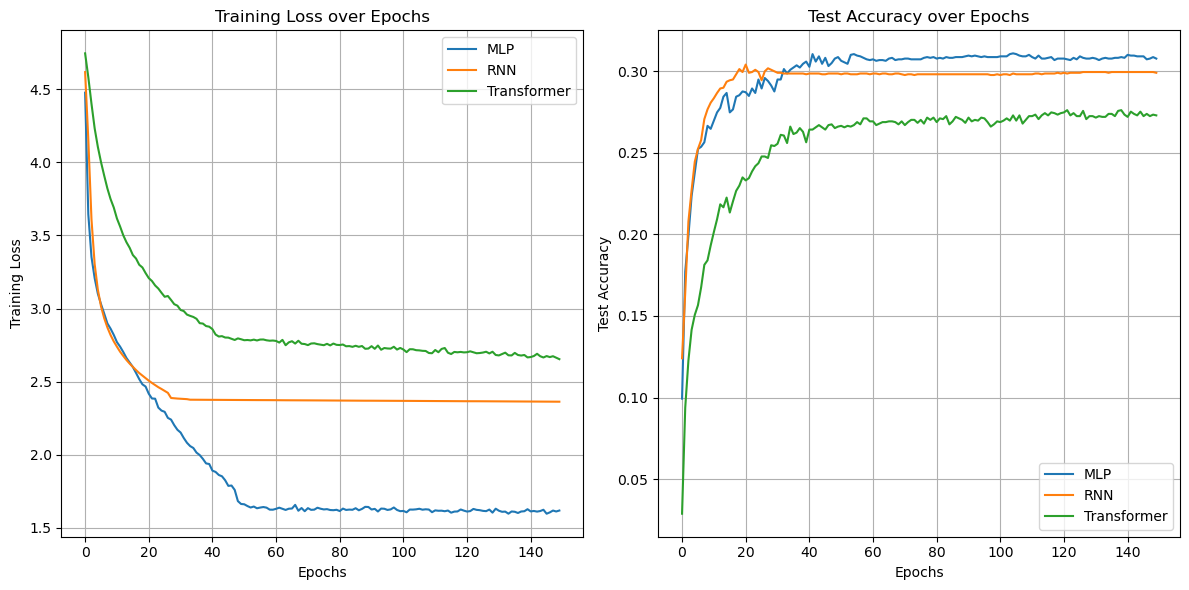}
  \caption{Training Loss (on the left) and Test Accuracy (on the right) employing TF-IDF for level-2 category.}
  \label{fig:tfidf2}
\end{figure}

After each epoch, the model is being evaluated on unseen data from the test set, where its predictions are compared to the actual labels to calculate accuracy. Both the training loss and test accuracy are recorded to monitor the model's progress. The figures below (Figure \ref{fig:glove1}, \ref{fig:tfidf1}, \ref{fig:glove2} and \ref{fig:tfidf2}) show the training process and the performance of the employed neural networks and Table \ref{tab:combined_results} shows their final test accuracy for both TF-IDF and GloVe. Discussion regarding the results will take place later, with all the available results being presented.

\begin{table}[h!]
\centering
\caption{Test Accuracy for both TF-IDF and GloVe after training}
\begin{tabular}{lcccc}
\toprule
\textbf{Classifier} & \multicolumn{2}{c}{\textbf{TF-IDF}} & \multicolumn{2}{c}{\textbf{GloVe}} \\
\cmidrule(lr){2-3} \cmidrule(lr){4-5}
& \textbf{level-1} & \textbf{level-2} & \textbf{level-1} & \textbf{level-2} \\
\midrule
MLP & 0.4945 & 0.3077 & 0.7138 & 0.4977 \\
RNN & 0.4675 & 0.2729 & 0.6841 & 0.4940 \\
TransformerEncoder & 0.4867 & 0.2990 & 0.7239 & 0.5302 \\
\bottomrule
\end{tabular}
\label{tab:combined_results}
\end{table}

\section{Machine Learning Models}
\label{sec:mlmodels}
It would be very helpful to have results from some Machine Learning (ML) models and therefore we decided to employ some of them. We are presenting the performance of three ML models: Support Vector Machine (SVM), Random Forest Classifier and Logistic Regression.

\subsection{Hyperparameter tuning}
Hyperparameter tuning involves optimizing the parameters of these three machine learning models to maximize their performance (classification accuracy) \cite{probst2019tunability}. A grid search methodology is employed, systematically evaluating various parameter combinations to identify the optimal configuration. We do not explicitly mention the parameters that being tested here, but further details can be found in the delivered source code (python notebooks). Instead, in Table \ref{tab:best-hyperparameters} we only present the best hyperparameters identified for each model. During the process of hyperparameter tuning, the performance for every model is assessed using 3-fold cross-validation, ensuring reliable parameter selection. 

\subsection{K-Fold cross validation}
K-Fold cross validation is a technique used to assess the performance of a machine learning model by dividing the dataset into k subsets or "folds." The model is trained and evaluated k times, each time using a different fold as the testing set and the remaining k-1 folds for training. This process helps ensure a more robust evaluation of the model's performance, as it considers multiple combinations of training and testing data. In our case, the final accuracy is the average of the accuracies computed in each iteration, providing a more reliable estimate of the model's generalization ability \cite{Anguita2012TheI}.

\begin{table*}[t] 
\centering
\caption{Best Hyperparameters for Each Model Across Categories and Methods after GridSearch}
\resizebox{\textwidth}{!}{ 
    \begin{tabular}{@{}ll@{}}
    \toprule
    \textbf{TFIDF - Category 1}& \textbf{GloVe - Category 1}\\ \midrule
    \begin{tabular}[c]{@{}l@{}}
    \textbf{Logistic Regression}: \{C: 1, max\_iter: 100, solver: lbfgs\} \\ 
    \textbf{Random Forest}: \{max\_depth: None, min\_samples\_split: 2, n\_estimators: 300\} \\ 
    \textbf{SVM}: \{C: 1, gamma: scale, kernel: rbf\} 
    \end{tabular} &
    \begin{tabular}[c]{@{}l@{}}
    \textbf{Logistic Regression}: \{C: 1, max\_iter: 100, solver: liblinear\} \\ 
    \textbf{Random Forest}: \{max\_depth: 20, min\_samples\_split: 2, n\_estimators: 300\} \\ 
    \textbf{SVM}: \{C: 10, gamma: scale, kernel: rbf\} 
    \end{tabular} \\ \midrule
    \textbf{TFIDF - Category 2}& \textbf{GloVe - Category 2}\\ \midrule
    \begin{tabular}[c]{@{}l@{}}
    \textbf{Logistic Regression}: \{C: 1, max\_iter: 100, solver: lbfgs\} \\ 
    \textbf{Random Forest}: \{max\_depth: None, min\_samples\_split: 5, n\_estimators: 300\} \\ 
    \textbf{SVM}: \{C: 1, gamma: scale, kernel: rbf\} 
    \end{tabular} & 
    \begin{tabular}[c]{@{}l@{}}
    \textbf{Logistic Regression}: \{C: 10, max\_iter: 100, solver: lbfgs\} \\ 
    \textbf{Random Forest}: \{max\_depth: None, min\_samples\_split: 2, n\_estimators: 300\} \\ 
    \textbf{SVM}: \{C: 10, gamma: scale, kernel: rbf\} 
    \end{tabular} \\ \bottomrule
    \end{tabular}
}
\label{tab:best-hyperparameters}
\end{table*}

\subsection{Machine Learning Results}
Having the best parameters for each model, we apply 5-fold cross-validation to better estimate their performance. Table \ref{tab:glove_results} and Table \ref{tab:tfidf_results} show the mean accuracy from the 5-fold cross validation for each model (the standard deviation in every experiment is almost 0.005). Discussion regarding the results will take place later, with all the available results being presented.
\begin{table}[H]
\centering
\caption{5-Fold Cross-Validation Results for GloVe}
\begin{tabular}{lcc}
\toprule
\textbf{Classifier} & \multicolumn{2}{c}{\textbf{Mean Accuracy}} \\
& \textbf{level-1} & \textbf{level-2} \\
\midrule
SVM Classifier & 0.7211 & 0.5570 \\
Random Forest Classifier & 0.6948 & 0.5072 \\
Logistic Regression Classifier & 0.6824 & 0.5585 \\
\bottomrule
\end{tabular}
\label{tab:glove_results}
\end{table}

\begin{table}[H]
\centering
\caption{5-Fold Cross-Validation Results for TF-IDF}
\begin{tabular}{lcc}
\toprule
\textbf{Classifier} & \multicolumn{2}{c}{\textbf{Mean Accuracy}} \\
& \textbf{level-1} & \textbf{level-2} \\
\midrule
SVM Classifier & 0.5008 & 0.3133 \\
Random Forest Classifier & 0.4934 & 0.2918 \\
Logistic Regression Classifier & 0.4640 & 0.2916 \\
\bottomrule
\end{tabular}
\label{tab:tfidf_results}
\end{table}

\section{Transfer Learning}
\label{sec:transferlearning}
Transfer learning is a machine learning technique where a model which is trained on one task is adapted to a different but quite similar task. The main idea is to apply the knowledge gained from a task with a lot of data to a task that has limited data. For instance, a model trained on a large dataset of general images can be fine-tuned to identify specific types of objects with only a small amount of labeled data. Most of the time, or always, this approach leads to better results compared to training a model from scratch. In the context of NLP, transfer learning has revolutionized the field. Pre-trained language models, such as BERT and GPT, are first trained on massive corpora to learn general language representations in order to capture the structure of the human language. These models are then fine-tuned on specific tasks, like sentiment analysis, text classification, or question answering, using much smaller datasets. 
\subsection{Pre-Trained Models}
Nowadays, there are many efficient language models available for use. In particular we test the efficacy of 7 different models: a) XLM-RoBERTa, b) DistilBERT, c) RoBERTa, d) BERT, e) ELECTRA, f) TinyBERT and g) ALBERT. Each of these models can be loaded with their \textbf{pre-trained} weights, allowing users to apply their capabilities out-of-the-box for a variety of NLP tasks:

\begin{enumerate}
    \item \textbf{XLM-RoBERTa \cite{conneau2019unsupervised}}: A multilingual model trained on 100 different languages. Unlike some XLM multilingual models, it does not require lang tensors to understand which language is used, and should be able to determine the correct language from the input ids.
    \item \textbf{DistilBERT \cite{sanh2019distilbert}}: A small, fast, cheap and light Transformer model trained by distilling BERT base. It has 40\% less parameters than google-bert/bert-base-uncased, runs 60\% faster while preserving over 95\% of BERT’s performances
    \item \textbf{RoBERTa \cite{liu2019roberta}}: An optimized variant of BERT, which modifies key hyperparameters, removing the next-sentence pre-training objective and training with much larger mini-batches and learning rates.
    \item \textbf{BERT \cite{kenton2019bert}}: The famous bidirectional transformer pre-trained using a combination of masked language modeling objective and next sentence prediction on a large corpus comprising the Toronto Book Corpus and Wikipedia.
    \item \textbf{ELECTRA \cite{clark2020electra}}: A pre-trained transformer model  with the use of another (small) masked language model. The inputs are corrupted by that language model, which takes an input text that is randomly masked and outputs a text in which ELECTRA has to predict which token is an original and which one has been replaced.
    \item \textbf{TinyBERT \cite{jiao2019tinybert}}: A model which is 7.5x smaller and 9.4x faster on inference than BERT-base and achieves competitive performances in the tasks of natural language understanding. It performs a novel transformer distillation at both the pre-training and task-specific learning stages.
    \item \textbf{ALBERT \cite{lan2019albert}}: A lightweight version of BERT with parameter reduction techniques like factorized embeddings, optimized for scalability without sacrificing accuracy.
\end{enumerate}

Pre-trained tokenizers are an essential component of transformer-based language models and they are responsible for converting raw text into numerical representations that can be processed by the model. These tokenizers are specifically trained to align with the vocabulary and tokenization strategy of their corresponding models, ensuring optimal performance and compatibility. Table \ref{tab:tokenizers} shows the models we used for our experiments, their corresponding tokenizers and the information for the pre-trained model weights.

\begin{table*}[!t]
\centering
\caption{The employed pre-trained models and tokenizers}
\begin{tabular}{lcc}
\toprule
\textbf{Model} & \textbf{Tokenizer} & \textbf{pre-trained from} \\
\midrule
AlbertForSequenceClassification & AlbertTokenizer & albert-base-v2\\
AutoModelForSequenceClassification & AutoTokenizer & huawei-noah\//TinyBERT\_General\_4L\_312D\\
ElectraForSequenceClassification & ElectraTokenizer & google\//electra-small-discriminator\\
BertForSequenceClassification & BertTokenizer & bert-base-uncased\\
RobertaForSequenceClassification & RobertaTokenizer & roberta-base\\
DistilBertForSequenceClassification & DistilBertTokenizer & distilbert-base-uncased\\
XLMRobertaForSequenceClassification & XLMRobertaTokenizer & xlm-roberta-base\\
\bottomrule
\end{tabular}
\label{tab:tokenizers}
\end{table*}

\subsection{PyTorch Data Format}
\begin{figure}[H]
    \centering
    \includegraphics[width=\linewidth]{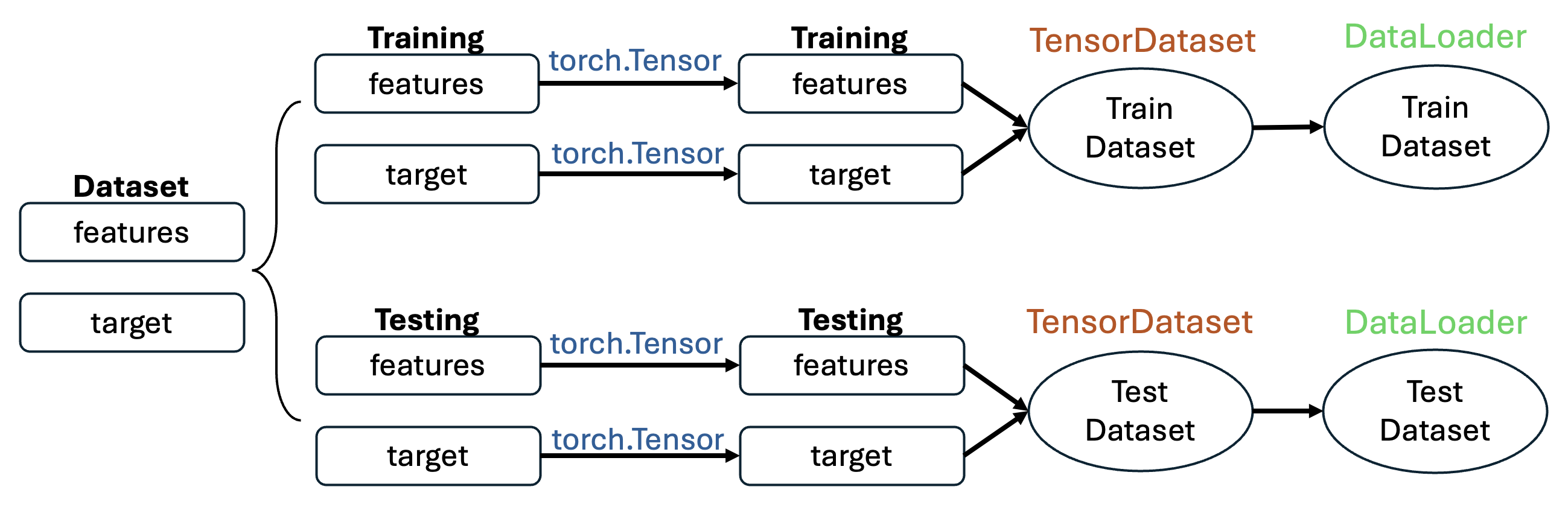}
    \caption{Pipeline for preparing the data for classifiers (neural networks).}
    \label{fig:dataset-pipeline}
\end{figure}

We split the dataset into training and testing sets and processes them into DataLoader objects, which are essential for efficient batch processing during training and evaluation. We convert the input data and labels into PyTorch tensors, then we generate attention masks to indicate non-padding tokens, and create TensorDataset objects. These datasets are then wrapped into DataLoaders with specified batch sizes (32), ensuring compatibility with transformer models. This function also shuffles the training data for better generalization and keeps the test data in order for a consistent evaluation. Figure \ref{fig:dataset-pipeline} demonstrates the whole process in our pipeline.

\subsection{Fine-tuning}
\begin{figure}[h]
    \centering
    \includegraphics[width=\linewidth]{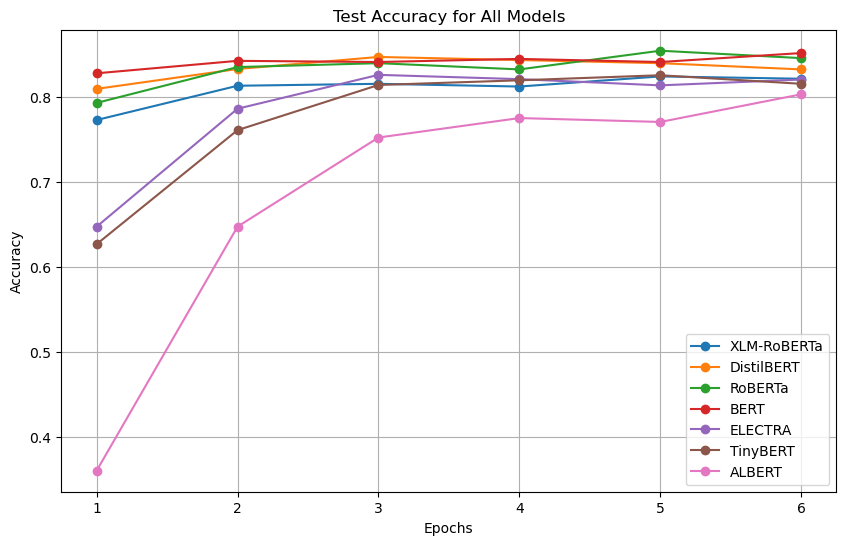}
    \caption{Test accuracy during training (6 epochs).}
    \label{fig:training-phase-accuracy}
\end{figure}

\begin{figure}[h]
    \centering
    \includegraphics[width=\linewidth]{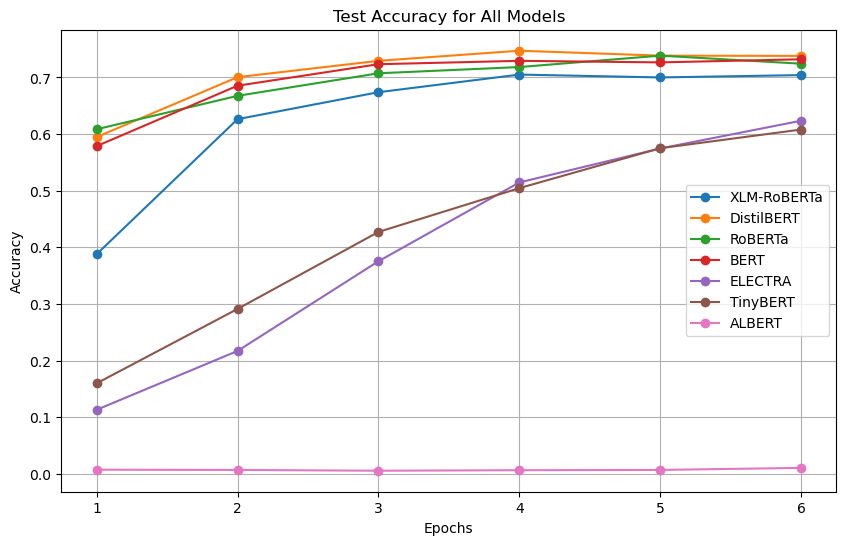}
    \caption{Test accuracy during training (6 epochs).}
    \label{fig:training-phase-accuracy-cat2}
\end{figure}

Once the data are ready and tokenized, we proceed to fine-tune the pre-trained models on our specific dataset. Fine-tuning involves adapting the pre-trained weights to the target task through additional training. Our training loop runs for a fixed number of six epochs, which we found to be sufficient for convergence in our experiments. During each epoch, the function tracks the training loss and evaluates the model's performance on a separate (unseen) test dataset, calculating accuracy. If the model achieves a significantly better accuracy during training, its weights are saved and we can restore them. At the end of training, the function restores these best weights for a final evaluation, ensuring that the model's performance is based on the most optimal parameters. We also record the time taken for the entire training process. This is especially useful for comparing the performance of different models during experiments. Figure \ref{fig:training-phase-accuracy} and Figure \ref{fig:training-phase-accuracy-cat2} demonstrate the performance during training (6 epochs) for each model for the two different categories - training loss is available in the delivered python notebook. It is clear that by the third or fourth epoch, the performance of most models stabilizes, indicating that additional training gives us minimal or not even better results. This highlights the efficiency and reliability of pre-trained transformer models for downstream tasks.

\subsection{Performance for Pre-Trained Models}
We train the model using the AdamW optimizer and CrossEntropy loss function, calculate the average training loss, and evaluate performance (accuracy) on the test (unseen) dataset after each epoch. A learning rate of 5e-5 is chosen, which is a common practice in fine-tuning, as it ensures gradual updates to model weights without significantly changing the pre-trained parameters during the 6 epochs of training. The 6 epochs that we apply are sufficient for fine-tuning as the models are pre-trained on large datasets, allowing them to learn general language representations that require minimal adjustment for the specific task. We save the model's state whenever a significant improvement is observed. After training, the best model weights are restored, ensuring optimal performance. The final accuracies after training are shown in Table \ref{tab:model_performance1} and Table \ref{tab:model_performance2} along with the size of each model (it terms of the number of parameters) and time taken for 6 epochs of fine tuning.

\begin{table}[h]
\centering
\caption{Performance of Pre-trained Models after Training for level-1 category}
\begin{tabular}{lccc}
\toprule
\makecell{\textbf{Model}} & \makecell{\textbf{No. of} \\ \textbf{Parameters}} & \makecell{\textbf{Final} \\ \textbf{Accuracy}} & \makecell{\textbf{Fine-Tuning} \\ \textbf{Time (min)}} \\
\midrule
XLM-RoBERTa & \textbf{278,056,721} & 0.8214 & \textbf{5m 17s}\\
DistilBERT & 66,966,545 & 0.8324 & 2m 28s\\
RoBERTa & 124,658,705 & 0.8457 & 4m 56s\\
BERT & 109,495,313 & \textbf{0.8516} & 4m 53s\\
ELECTRA & 13,553,169 & 0.8205 & 1m 28s\\
TinyBERT & 14,355,569 & 0.8155 & 0m 46s\\
ALBERT & 11,696,657 & 0.8031 & 5m 10s\\
\bottomrule
\end{tabular}
\label{tab:model_performance1}
\end{table}

\begin{table}[h]
\centering
\caption{Performance of Pre-trained Models after Training for Level-2 Category}
\begin{tabular}{lccc}
\toprule
\makecell{\textbf{Model}} & \makecell{\textbf{No. of} \\ \textbf{Parameters}} & \makecell{\textbf{Final} \\ \textbf{Accuracy}} & \makecell{\textbf{Fine-Tuning} \\ \textbf{Time (min)}} \\
\midrule
XLM-RoBERTa & \textbf{278,127,469} & 0.7042 & \textbf{5m 17s} \\
DistilBERT & 67,037,293 & \textbf{0.7381} & 2m 28s \\
RoBERTa & 124,729,453 & 0.7244 & 4m 56s \\
BERT & 109,566,061 & 0.7321 & 4m 53s \\
ELECTRA & 13,576,813 & 0.6236 & 1m 27s \\
TinyBERT & 14,384,365 & 0.6081 & 0m 45s \\
ALBERT & 11,767,405 & 0.0105 & 5m 8s \\
\bottomrule
\end{tabular}
\label{tab:model_performance2}
\end{table}

\section{Results}
\label{sec:results}
We already presented how we trained and evaluated all the models and in this section we are presenting the final performance (in terms of accuracy) of each model that we experimented with. Table \ref{tab:final_results1} and Table \ref{tab:final_results2} clearly demonstrate the results for both level-1 and level-2 categories. Later, in Section \ref{sec:discussion} we discuss about the results.
\begin{table}[h]
\centering
\caption{Final Results for Level-1 Category}
\begin{tabular}{lcc}
\toprule
\textbf{Model} & \multicolumn{2}{c}{\textbf{Accuracy}} \\ 
\midrule
XLM-RoBERTa & \multicolumn{2}{c}{0.8214} \\
DistilBERT & \multicolumn{2}{c}{0.8324} \\
RoBERTa & \multicolumn{2}{c}{0.8457} \\
\textbf{BERT} & \multicolumn{2}{c}{\textbf{0.8516}} \\
ELECTRA & \multicolumn{2}{c}{0.8205} \\
TinyBERT & \multicolumn{2}{c}{0.8155} \\
ALBERT & \multicolumn{2}{c}{0.8031} \\
\midrule
 & \textbf{Accuracy (TFIDF)} & \textbf{Accuracy (GloVe)} \\ 
\midrule
MLP & 0.4945 & 0.7138 \\
RNN & 0.4675 & 0.6841 \\
TransformerEncoder & 0.4867 & \textbf{0.7239}  \\
\midrule
SVM & \textbf{0.5008} & \textbf{0.7211} \\
Random Forest & 0.4934 & 0.6948 \\
Logistic Regression & 0.4640 & 0.6824 \\
\bottomrule
\end{tabular}
\label{tab:final_results1}
\end{table}

\begin{table}[h]
\centering
\caption{Final Results for Level-2 Category}
\begin{tabular}{lcc}
\toprule
\textbf{Model} & \multicolumn{2}{c}{\textbf{Accuracy}} \\ 
\midrule
XLM-RoBERTa & \multicolumn{2}{c}{0.7042} \\
DistilBERT & \multicolumn{2}{c}{\textbf{0.7381}} \\
RoBERTa & \multicolumn{2}{c}{0.7244} \\
BERT & \multicolumn{2}{c}{\textbf{0.7321}} \\
ELECTRA & \multicolumn{2}{c}{0.6236} \\
TinyBERT & \multicolumn{2}{c}{0.6081} \\
ALBERT & \multicolumn{2}{c}{0.010} \\
\midrule
 & \textbf{Accuracy (TFIDF)} & \textbf{Accuracy (GloVe)} \\ 
\midrule
MLP  & 0.3077 & 0.4977 \\
RNN  & 0.2729  & 0.4940 \\
TransformerEncoder & 0.2990 & 0.5302 \\
\midrule
SVM & \textbf{0.3133} & \textbf{0.5570} \\
Random Forest & 0.2918 & 0.5072 \\
Logistic Regression & 0.2916 & \textbf{0.5585} \\
\bottomrule
\end{tabular}
\label{tab:final_results2}
\end{table}

\section{Discussion}
\label{sec:discussion}
The results indicate that pre-trained transformer models outperform traditional models across both categories. Among pre-trained models, \textbf{BERT} achieved the highest accuracy (0.8516) in the level-1 category and demonstrates its strong capability to capture contextual information. Similarly, in the level-2 category, \textbf{BERT} and \textbf{DistilBERT} showed comparable performance, achieving accuracies of 0.7321 and 0.7381, respectively. Compared to other neural networks and machine learning models, these results highlight the robustness of pre-trained models in handling such classification tasks.

However, we observe a performance gap between level-1 and level-2 tasks for all models, with lower accuracy in the latter. Certrainly, this is because of the increased complexity (107 classes instead of 17 in level-1), which means that this is a more difficult challenge for both pre-trained and traditional models.

Traditional machine learning models showed reasonable performance when paired with TFIDF and GloVe embeddings. Notably, the combination of GloVe embeddings with these models resulted in improved accuracies, suggesting that pre-trained word embeddings can enhance the effectiveness of traditional models. For example, SVM achieved accuracy=0.7211 on level-1 tasks with GloVe embeddings, significantly higher than its performance with TF-IDF where it achieved accuracy=0.5008. However, these models consistently underperformed compared to the pre-trained ones, showing the limitations of traditional approaches. We observed a very bad performance from the ALBERT model in level-2 category where it achieved just 0.01 accuracy which is something very strange taking into consideration its performance in level-1 category. One possible reason is that it probably needs more epochs during fine-tuning. However, we didn't have the GPU capacity to experiment more on this and we kept everything consistent at 6 epochs.

\subsection{Key Observations}
\begin{itemize}
    \item \textbf{Number of Parameters for Pre-trained Models:} While models like TinyBERT and ALBERT offer efficiency due to their smaller sizes, their performance was lower compared to larger models such as BERT and RoBERTa. This highlights a trade-off between model size and performance, particularly in complex classification tasks (such as level-2 category).
    \item \textbf{Embedding Choice Matters:} The results for TFIDF and GloVe embeddings show the importance of feature representation in traditional models. GloVe embeddings, with their pre-trained information, consistently outperformed TFIDF across both level-1 and level-2 categories.
\end{itemize}


\section{Conclusions}
\label{sec:conclusions}
In this work, we evaluated the performance of pre-trained transformer models and traditional machine learning approaches on two classification tasks (level-1 and level-2). Our findings demonstrate that pre-trained models, such as \textbf{BERT} and \textbf{RoBERTa}, consistently outperform traditional models. However, we observed a performance decrease for all models on the more complex task (level-2). Additionally, traditional models paired with GloVe (pre-trained) embeddings showed competitive performance in simpler tasks, pointing out the importance of embedding quality in feature-based approaches. It is also noteworthy to say, that if we do not care about accuracy and correct predictions in general and we have computational limitations, it is better to go with a standard machine learning model (with GloVe) which is significantly smaller and easier to deploy compared to pre-trained ones. Overall, the findings emphasize the dominance of pre-trained transformer-based architectures for text classification tasks.



\end{document}